\documentclass[conference]{IEEEtran}

\IEEEoverridecommandlockouts                              

\usepackage{times}
\usepackage[numbers]{natbib}
\usepackage{multicol}
\usepackage[bookmarks=true]{hyperref}
\usepackage{amsmath,amssymb,amsfonts}
\usepackage{algorithmic}
\usepackage{graphicx}
\usepackage{textcomp}
\usepackage{verbatim}
\usepackage[linesnumbered,ruled,vlined]{algorithm2e}
\usepackage{xcolor}
\usepackage{subcaption}
\usepackage{multirow}

\newcommand{\locc}{{\sc locc}}
\newcommand{\loccsim}{{\sc brax-locc}}
\newcommand{\igsphere}{{\sc ig-sphere}}
\newcommand{\braxsphere}{{\sc brax-sphere}}
\newcommand{\igcvxd}{{\sc ig-cvxd}}
\newcommand{\igcvxh}{{\sc ig-cvxh}}
\newcommand{\iscd}{{\sc is-cd}}
\newcommand{\ucfgjk}{{\sc ucf-gjk}}
\newcommand{\ocnglobal}{{\sc global-ocn}}

\title{
Local object crop collision network for efficient simulation of non-convex objects
in GPU-based simulators
}

\author{Dongwon Son and Beomjoon Kim%
\thanks{This work was supported by Institute of Information \& communications Technology Planning \& Evaluation (IITP) grant funded by the Korea government(MSIT)  (No.2019-0-00075, Artificial Intelligence Graduate School Program(KAIST)) and Institute of Information \& communications Technology Planning \& Evaluation (IITP) grant funded by the Korea government(MSIT) (No.2022-0-00311, Development of Goal-Oriented Reinforcement Learning Techniques for Contact-Rich Robotic Manipulation of Everyday Objects)}%
\thanks{The authors are with Graduate School of AI, KAIST, Seoul, Republic of Korea
{\tt\small \{dongwon.son beomjoon.kim\}@kaist.ac.kr}}%
}

\begin{document}

\maketitle
\thispagestyle{empty}
\pagestyle{empty}

\begin{abstract} Our goal is to develop an efficient contact detection algorithm for large-scale GPU-based simulation of non-convex objects. Current GPU-based simulators such as IsaacGym~\cite{makoviychuk2021isaac} and Brax~\cite{freeman2021brax} must trade-off speed with fidelity, generality, or both when simulating non-convex objects. Their main issue lies in contact detection (CD): existing CD algorithms, such as Gilbert–Johnson–Keerthi (GJK), must trade off their computational speed with accuracy which becomes expensive as the number of collisions among non-convex objects increases. We propose a data-driven approach for CD, whose accuracy depends only on the quality and quantity of offline dataset rather than online computation time. Unlike GJK, our method inherently has a uniform computational flow, which facilitates efficient GPU usage based on advanced compilers such as XLA (Accelerated Linear Algebra)~\cite{xla}. Further, we offer a data-efficient solution by learning the patterns of colliding local crop object shapes, rather than global object shapes which are harder to learn. We demonstrate our approach improves the efficiency of existing CD methods by a factor of 5-10 for non-convex objects with comparable accuracy. Using the previous work on contact resolution for a neural-network-based contact detector~\cite{son2020sim}, we integrate our CD algorithm into the open-source GPU-based simulator, Brax, and show that we can improve the efficiency over IsaacGym and generality over standard Brax. We highly recommend the videos of our simulator included in the supplementary materials. \url{https://sites.google.com/view/locc-rss2023/home}

\end{abstract}

\IEEEpeerreviewmaketitle

\section{Introduction}

With an ever-increasing demand for bigger datasets, GPU-based simulators are becoming an essential tool in robotics. Unlike CPU-based simulators, they can simulate thousands of environments in parallel, which makes big data generation extremely efficient. In fact, several works in robot manipulation and locomotion have empirically demonstrated that by utilizing GPU-based simulators, you attain a significant training speed advantage over CPU-based simulators \cite{makoviychuk2021isaac, freeman2021brax}.

\begin{figure}[htb]
    \centering
    \includegraphics[width=0.48\textwidth]{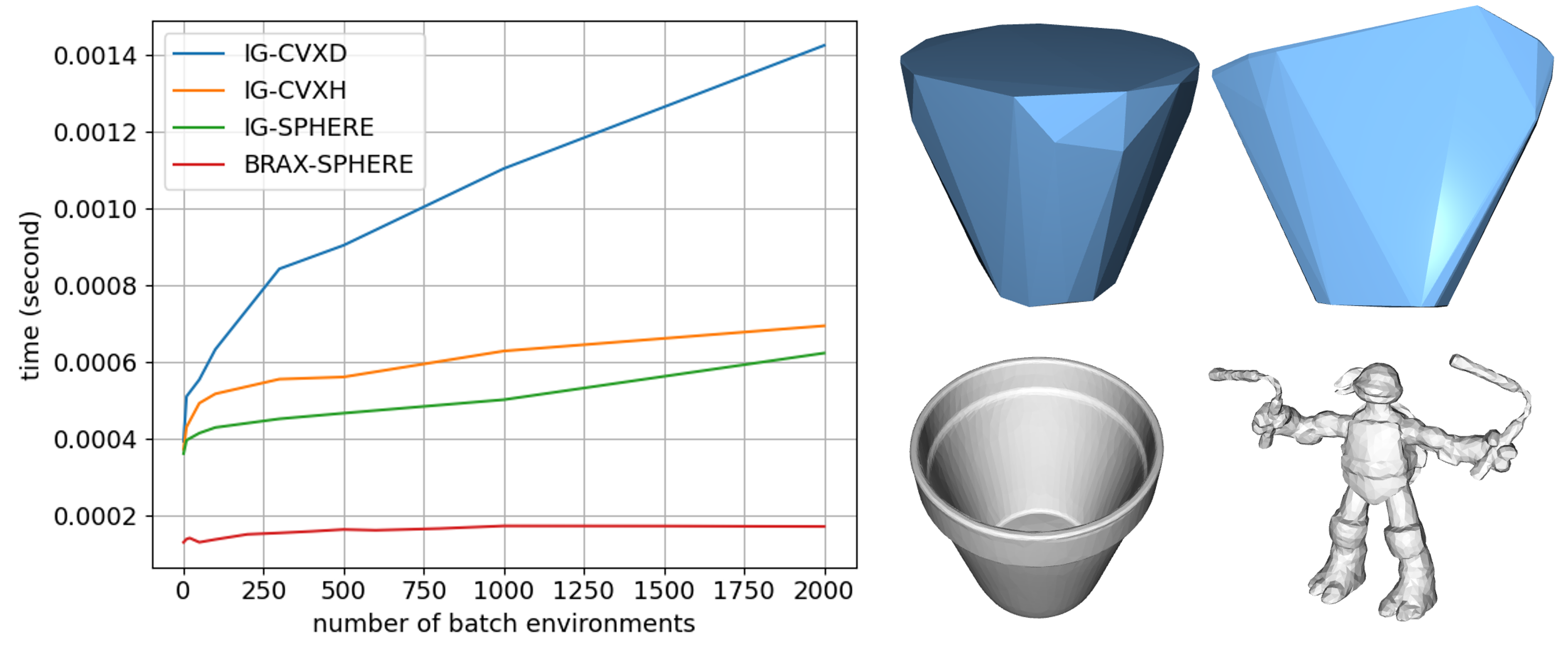}
    \caption{(Left) Time for simulating a single time step vs. the number of environments using IsaacGym and Brax in object dropping tasks. (Right) An example of the convex hull approximation of objects used by~\igcvxh. 
    }
    \label{fig:IG_and_Brax_speed_plot}
\end{figure}

While parallelism improves efficiency over CPU-based simulators, the current state-of-the-art GPU-based simulators, like IsaacGym~\cite{makoviychuk2021isaac}, slow down significantly when simulating non-convex objects as the number of environments grows.  This is well demonstrated in Figure~\ref{fig:IG_and_Brax_speed_plot} (left). The simulation time of \igsphere, which simulates dropping spheres in IsaacGym, remains more or less similar even as we increase the number of environments. However, when you simulate \emph{non}-convex objects, the speed starts to deteriorate rapidly with respect to the number of environments, as demonstrated by \igcvxd, which approximates non-convex objects using convex decomposition~\cite{vhacd}.

\begin{figure}
    \centering
    \includegraphics[width=0.48\textwidth]{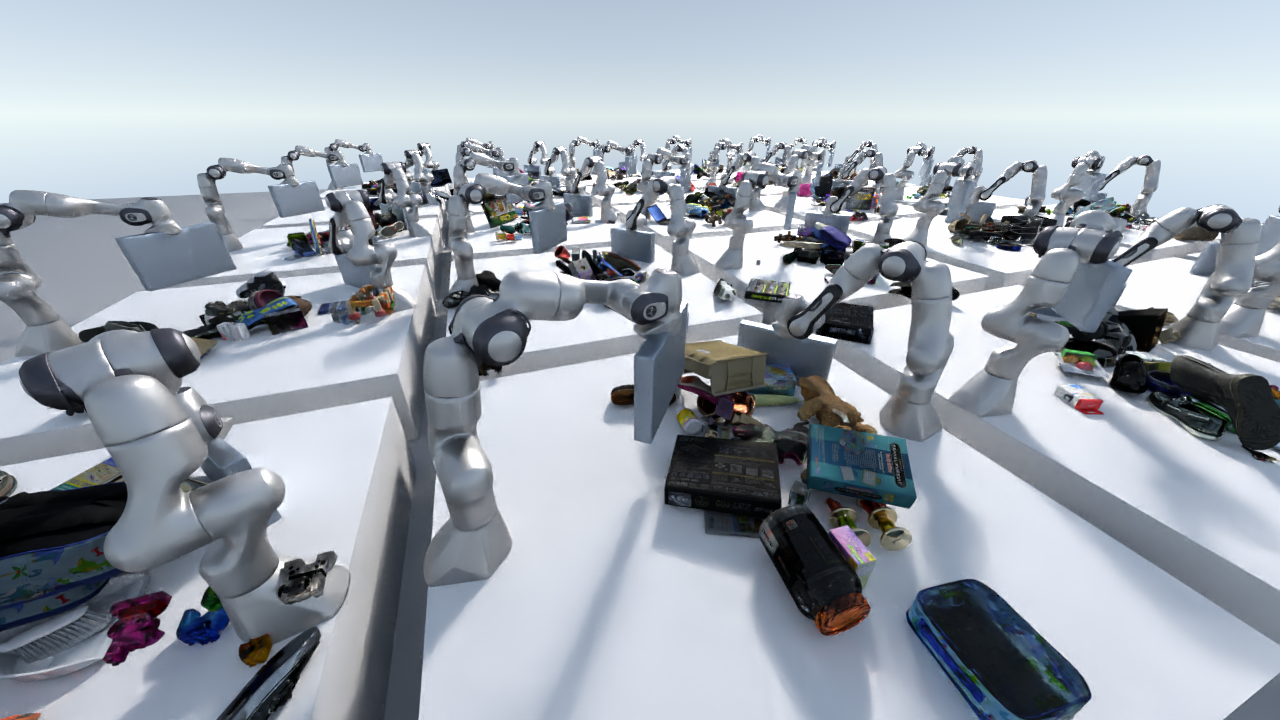}
    \caption{Snapshot of parallel simulation with \loccsim, where two Franka Emika PANDA robots are pushing 25 objects most of which are non-convex. Qualitatively, it robustly simulates heavy interactions between non-convex objects. We recommend the readers check the video included in the supplementary material.}
    \label{fig:push_demo}
\end{figure}

The fundamental reason for the slowdown in IsaacGym is CD. The most widely-used CD algorithm, GJK~\cite{gilbert1988fast}, only works with convex objects and needs convex decomposition for non-convex objects. So to improve its accuracy, we need to increase the number of elements in the decomposition, but this significantly increases online computational time due to an increase in the pairs of elements for which we need to check the collisions.

Alternatively, we can use a convex hull approximation of the non-convex objects, denoted \igcvxh, whose speed also remains more or less constant with respect to the number of environments as shown in Figure~\ref{fig:IG_and_Brax_speed_plot} (left). However, the degradation in simulation fidelity is typically too high to bear, as shown in the approximated collision mesh of objects in Figure~\ref{fig:IG_and_Brax_speed_plot} (right).

Furthermore, GJK involves branching, such as if-statements, and does not have a uniform computational flow (UCF). As a result, it is difficult to benefit from the advanced optimization techniques available in domain-specific GPU compilers such as XLA~\cite{xla}. This is well illustrated in the comparison of IsaacGym to Brax~\cite{braxblog}. Brax is a GPU-based simulator that, unlike IsaacGym, has a UCF for all of its computations. When we compare Brax and IsaacGym in simulating dropping spheres, Figure~\ref{fig:IG_and_Brax_speed_plot} (left) shows that
the speed of \braxsphere~is much higher than \igsphere. However, this comes at a cost in generality: Brax is limited to simulating only convex objects, due to the lack of contact detection algorithms that have a UCF \emph{and} handle general shapes. Furthermore, Brax
requires all environments to have the same set of objects to maintain its UCF.

In~\cite{son2020sim}, the authors created a contact resolution algorithm for a neural network (NN)-based contact detector. They demonstrated that their method is highly accurate and reliable, even when simulating a complex task of tightening a nut onto a bolt. However, their contact detector has a limitation in that it cannot generalize to different shapes. Our goal is to create a NN-based contact detector that can handle a variety of shapes and can easily be used within a physics engine with~\cite{son2020sim} as a contact resolution algorithm. Our approach has a performance advantage because NNs do not have branching in their computations, allowing for the use of XLA. Additionally, compared to GJK which needs to increase its online computation time for better accuracy, the accuracy of our NN-based contact detector only depends on the quality and quantity of offline collected data, while the online computation time remains constant as a NN prediction.

One simple idea to implement an NN-based collision detector is to adapt SceneCollisionNet~\cite{danielczuk2021object} that predicts a collision between an object and the scene. That is, we first sample a point cloud from an object mesh, encode each point cloud into a shape embedding using a shape encoder, and then use the shape embeddings to predict a collision. However, this turns out to be data inefficient: because it requires the network to learn to capture the entire object shape, if we wish to generalize to a variety of objects, we would need a large number of realistic object mesh models which are expensive to obtain.

\begin{figure}[htb]
    \centering
    \includegraphics[width=0.45\textwidth]{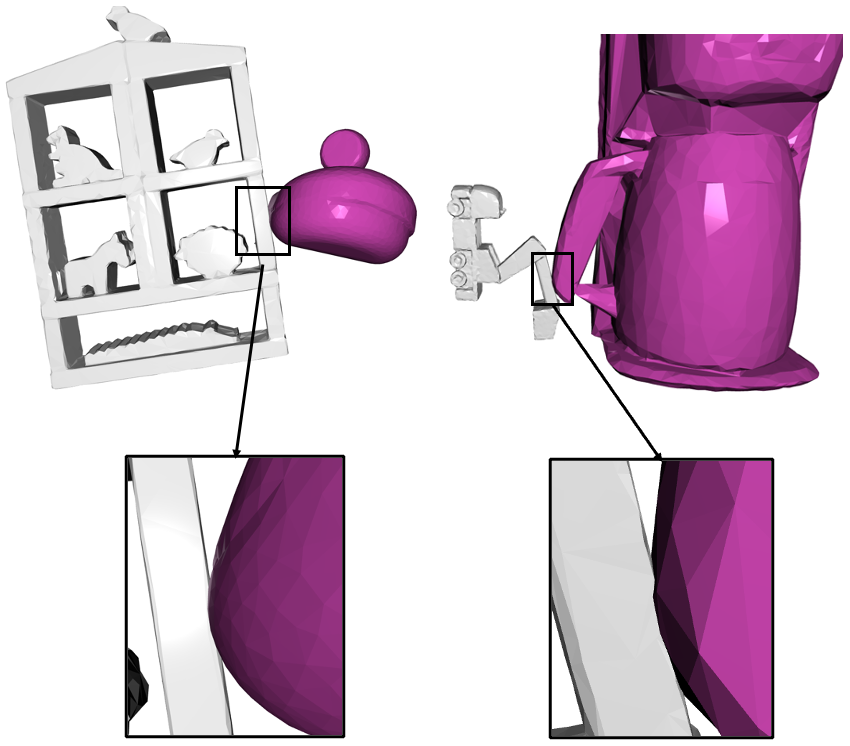}
    \caption{
    Consider the figures on the top row, where two different pairs of objects are in collision. To accurately predict a collision, a naive object collision detector would need to first learn to represent the shapes of objects, but this requires many object shape data because there is a large variability in object shapes, especially for non-convex objects. Now consider the figures on the bottom row. In the local crops where collisions occur, we see that their shapes are extremely similar
    even though the global object shapes are totally different. 
    Such similarity in local crops of object shapes is what we try to exploit in our algorithm to improve data efficiency.
    }
    \label{fig:locality_intuition}
\end{figure}

Inspired by~\cite{jiang2020local,chabra2020deep}, we instead propose to encode only the shape of local crops of two objects that are in collision. Our intuition is that the patterns in local crops of object shapes are easier to learn than the patterns in global object shapes, as they are more frequent in data, and can be generated more cheaply. This intuition is demonstrated in Figure~\ref{fig:locality_intuition}. 

\begin{figure*}[htb]
\centering
\includegraphics[width=0.95\textwidth]{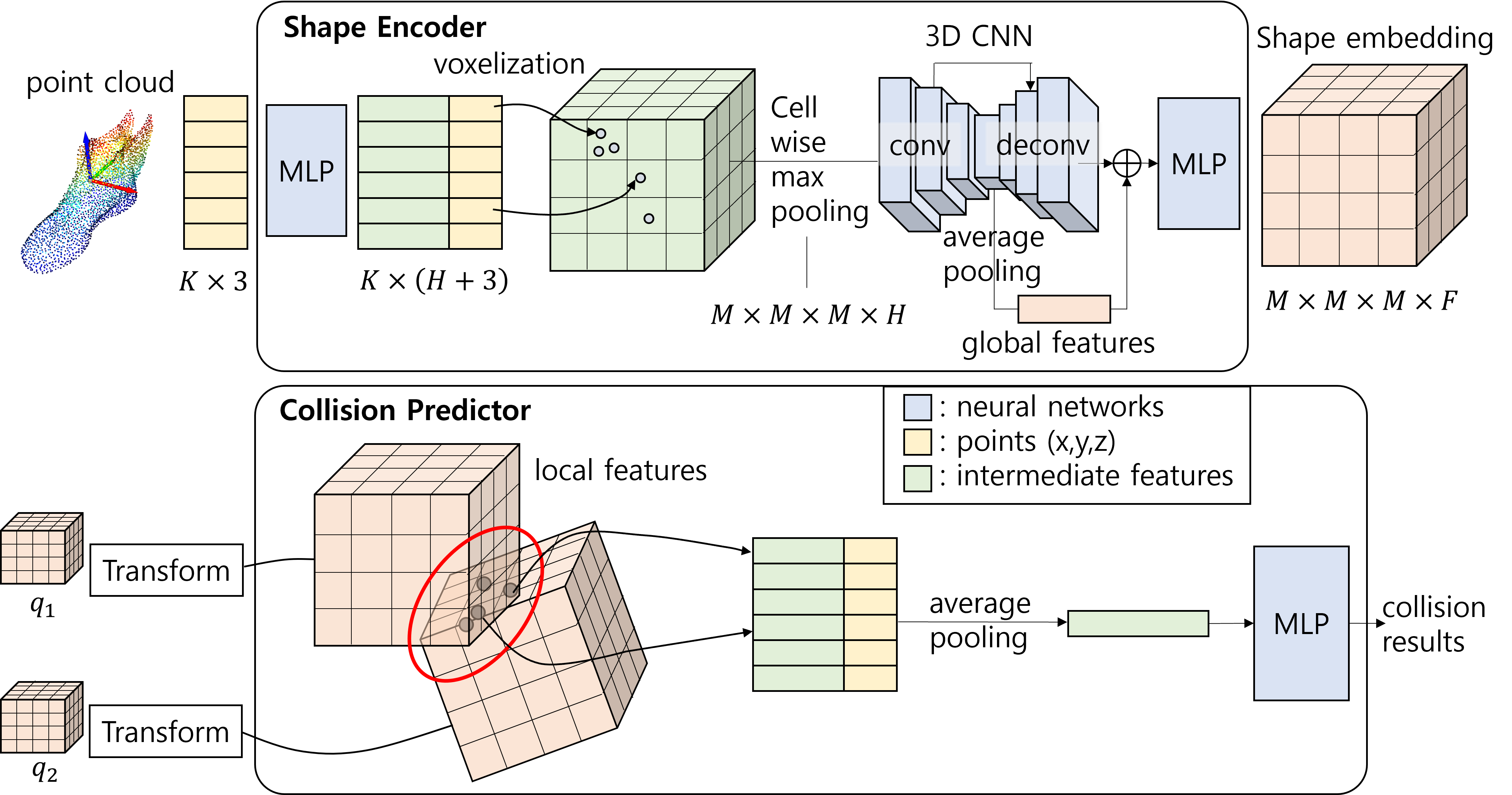}
\caption{
Computational flow in \locc. In \locc, there are two modules: shape encoder, and collision predictor. The shape encoder takes an object shape represented in point cloud as an input, and generates a shape embedding. The collision predictor takes two shape embeddings and object poses as inputs, and outputs a collision result.  $K$
denotes the number of points in the point cloud of an object shape, $M$ denotes the size of the voxel grid defined on the AABB of an object shape, and $H$ and $F$ denote the output dimensions of the first and
second MLPs in the shape encoder respectively. Cross-circle symbol in the shape encoder indicates concatenation.}
\label{fig:LOCC_architecture}
\end{figure*}

Our algorithm, called \locc~(Local Object Crop Collision Network), implements this idea
by creating local features of object shapes. More concretely, given the poses and shapes of two objects of interest, we first define the voxel grid on the Axis-Aligned Bounding Box (AABB) each object. Then, we use a shape encoder to compute a feature for each cell of the voxel grid, and use the poses to create Oriented Bounding Box (OBB) of the voxel grid of features. We check which cells are in collision, and then pass only the features from the colliding cells to a collision predictor. Figure~\ref{fig:LOCC_architecture} demonstrates the computations in our \locc.

In our experiments, we first demonstrate that \locc~outperforms state-of-the-art NN-based collision detectors, such as \ocnglobal, an adaptation of~\cite{danielczuk2021object}, in terms of speed and data efficiency. Second, we implement a variant of GJK that has a UCF and show that \locc~has a 5-10 times speed gain while having a comparable accuracy. Finally, we extend Brax, the open-source GPU-based simulator, with \locc~and the contact resolution algorithm from~\cite{son2020sim} to support non-convex objects. We show that the resulting simulator, \loccsim, can simulate non-convex objects at a higher speed than IsaacGym and more general objects than standard Brax. An example scene from \loccsim~is shown in Figure~\ref{fig:push_demo}.

\section{Related Works}



\textbf{Collision detection using a neural network}
Recently, several methods have been proposed for \emph{neural implicit functions}, which use a neural network to represent a shape. Given spatial information such as locations in 3D space, neural implicit functions output values that represent shapes~\cite{park2019deepsdf, jiang2020local, mueller2022instant, takikawa2021neural}. Based on these advances, several methods have been proposed for collision detection using neural implicit functions where a a set of query points is used to test if a query point lies inside of both shapes~\cite{xu2021end, driess2022learning}. However, these methods typically require a large number of query points because their accuracy depends heavily on the density of the query points. Further, explicitly reconstructing the shapes of objects, which requires a large amount of expensive shape data, has shown to be unnecessary for contact detection~\cite{danielczuk2021object}. 

So instead, several methods~\cite{das2020learning, danielczuk2021object} propose to predict collision directly to improve data efficiency and speed. More concretely, in~\cite{danielczuk2021object}, the authors proposed a method that, given a point cloud of an object of interest, its pose, and a point cloud of a scene, determines if there is a collision between the object and the scene.  In~\cite{das2020learning}, the authors propose a method for learning a function for representing the configuration space obstacles which is used to check collision for a given robot configuration. Both of these approaches require learning the representation of the shape of the entire scene or object. In contrast, we learn a representation of a 
colliding local crops of objects to improve data efficiency.

Further, the primary purpose of~\cite{das2020learning, danielczuk2021object}~is to improve the motion planning speed. For this, it is sufficient to determine if a collision has occurred in the current world state. However, for simulation, we not only need to detect the collisions but also determine between which objects the contact has occurred to perform contact resolution. So, we need an \emph{object} collision detector that checks the collision between a pair of objects, rather than between a scene and an object or robot. We will refer to a NN that takes shapes and poses of
two objects of interest as inputs and predicts a collision as \emph{Object Collision Net} (OCN).

\textbf{Collision resolution using an OCN}
The goal of contact resolution is to prevent penetration and resolve contacts so that the colliding objects move in a direction that is in accordance with the physics law after contact. In~\cite{son2020sim}, the authors have proposed the method for contact resolution using an OCN. Since we use this method without modification when we integrate~\locc~into Brax, we describe briefly describe how it works.

For contact resolution, we need two quantities: the set of contact points and contact forces. To find the contact points using an OCN, we first use the OCN to detect at what object pose the collision occurs. Then, we determine the directions of contact force and torque at object's center of mass by computing the gradient of the OCN with respect to the pose. The intuition is that the steepest pose direction that gets the objects out of collision is given by the gradient of the OCN, which must be in the same direction as the contact force and torque. Using these force and torque directions, we then find the contact points using the Point Isolation method~\cite{haddadinTRO2017}.

Once the contact points have been determined, we define the contact constraints that enforces the colliding objects to move in the direction of not penetrating further
at the contact point. The contact forces are computed using trajectory optimization subject to contact constraints and spring motion constraints based on penetration depth \cite{moore1988collision, anitescu1997formulating}. Based on these quantities, the objects follow a simple rigid body dynamics. For more details, such as how to deal with contact resolution at the equilibrium, we refer the readers to the original paper~\cite{son2020sim}.

While \cite{son2020sim} has demonstrated the effectiveness of their contact resolution method in the challenging task of screwing a nut into a bolt, their OCN was limited to a single bolt and nut and cannot take object shapes as inputs. Our algorithm, \locc, can be seen as a generalization of~\cite{son2020sim} to a variety of shapes.

\textbf{Analytical collision detection methods and their usage in GPUs}
GJK \cite{gilbert1988fast} is used in a variety of representative physics engines, such as Havok, PhysX, and Bullet. In practice, GJK can detect collisions typically in constant time for convex objects. However, for non-convex objects, we must first make a convex decomposition of all the objects using a method such as V-HACD~\cite{mamou2016volumetric}, run GJK to check collision for each pair of decomposition elements in the worst case\footnote{If objects are sufficiently faraway, you can rule them out in the Broad phase of collision detection.}, then report collision if any one of the element pairs is in a collision.

Since GPUs expedite computations by applying the same function across different environments, it is hard to efficiently use GJK on a GPU due to branching in computations. We can, as we show in our experiments, modify GJK to have a UCF. However, even in this case, the computational cost of traversing through the pairs of convex elements outweighs that of the simple feed-forward prediction in~\locc.

Separate Axis Theorem (SAT) is another widely used method for detecting collisions between convex objects. It works by projecting the shapes onto different axes and checking for overlap in all dimensions. If there is no overlap on any axis, then the objects do not collide. This method can be efficiently implemented on GPUs, as each axis can be processed in parallel and the results can be combined to determine if a collision has occurred. SAT requires only simple arithmetic operations, such as dot products and comparisons, which can be easily accelerated by GPUs and is used in Brax. Despite its efficiency, SAT has limited scaling capability when dealing with complex non-convex shapes because it requires checking over $N_1N_2$ axes, where $N_1$ and $N_2$ are the numbers of edges for objects 1 and 2 respectively.

\section{Local object crop collision network}

\label{sec:local_object_crop_collision_network}
We now describe \locc~which directly evaluates the collision between two objects using their meshes and poses. There are two modules in \locc: shape encoder and collision predictor. Like SceneCollisionNet~\cite{danielczuk2021object}, our shape encoder uses a voxel grid defined on the object point cloud and computes cell-wise features. Unlike SceneCollisionNet however, we exploit locality by passing to the collision predictor only the local crops of the shape embeddings that are in a collision to facilitate data efficiency. We now walk through Figure~\ref{fig:LOCC_architecture} that describes the computations in \locc.


We first create a point cloud by sampling $K$ number of points using the mesh of an object. The shape encoder begins its computation by processing each point using a multilayer perceptron (MLP), which outputs $K\times H$ features, where $H$ is the output dimension of the MLP. We then compute the AABB of each object's mesh and define a voxel grid  of size $M\times M\times M$ over the AABB. We put the features into the voxel grid according to the coordinates of their corresponding points to get features of size $M\times M \times M\times H$. We process the feature for each cell using cell-wise max-pooling over the features. We further process the local feature for each cell by applying  a 3D Convolutional Neural Networks (CNNs) with skip connections to generate the shape embedding of size $M \times M\times M \times F$, where $F$ is the dimension of the feature at each cell. 

In the 3D CNN, we use a U-Net-like architecture~\cite{ronneberger2015u}~but add an average pooling operation just before the deconvolution to extract a global feature. This global feature is then broadcasted to the local features of each cell just after the deconvolution. The intuition is to encode both local and global features into each cell feature. 

For the collision predictor, we first create the OBB of shape embeddings of two objects by using their poses. Then, we select the features in colliding cells. One way to obtain the cells in a collision is by considering the center point of each cell of the OBB, and checking if that point is inside of another object's OBB. The downside of this approach is that while the center point may be outside the OBB, the cells may still be in a collision. To solve this, we add a margin to each cell of OBB whose magnitude is the distance from the center point to a vertex of a cell. This way, we are guaranteed not to have false negatives in detecting the cells in a collision. Using the features of the selected cells, we apply average pooling, and the resulting vector gets passed to a collision predictor which then makes a prediction. 

Note that while margin-padding guarantees no false negative in collision detection, it may result in false positives, and include more features than needed. However, we found that adding more information does not usually hurt the performance of collision prediction.


There are some notable hyperparameters that are worth mentioning. First, $M$ is a particularly important hyperparameter and should be chosen based on object sizes.
If $M$ is too small, then the shape embedding will not contain sufficient information. If $M$ is too big, then it will take up a lot of memory and checking collisions between two OBBs would take a long time. After testing values of 5, 6, and 7, we found that 6 provides a good balance of accuracy and efficiency.

For $K$, which is the number of points fed into the shape encoder, its value doesn't impact testing efficiency because we keep the shape embedding at a fixed size through cell-wise max pooling, even when $K$ changes. We set $K$ to 1500 so that we can generate enough points to cover the surface and capture shape details.



\subsection{Dataset preparation and training}
\label{sec:dataset_preparation_and_training}
To prepare the dataset we use existing object datasets such as YCB or Google ScanNet~\cite{calli2015ycb, 
downs2022google}. Our dataset has the form $\{(x^{(i)}_1, x^{(i)}_2, q^{(i)}_1, q^{(i)}_2, y^{(i)})\}_{i=1}^{n}$ 
where $x_{1}$ and $x_{2}$ denote object meshes, 
$q_1$ and $q_2$ denote object poses, and $y$ denotes a collision label. 
This data is generated by synthetically creating a collision between two objects. 

A naive approach to generate such a dataset is to first sample two objects from an object set, place them at poses uniformly sampled from a pre-defined bound, and evaluate the collision to assign $y$. However, we found that  uniform random sampling mostly generates trivial cases where two objects are either too far away or overlap severely. This would make~\locc~fragile to non-trivial cases where objects overlap only slightly.

\begin{figure}[h]
    \centering
    \includegraphics[width=0.35\textwidth]{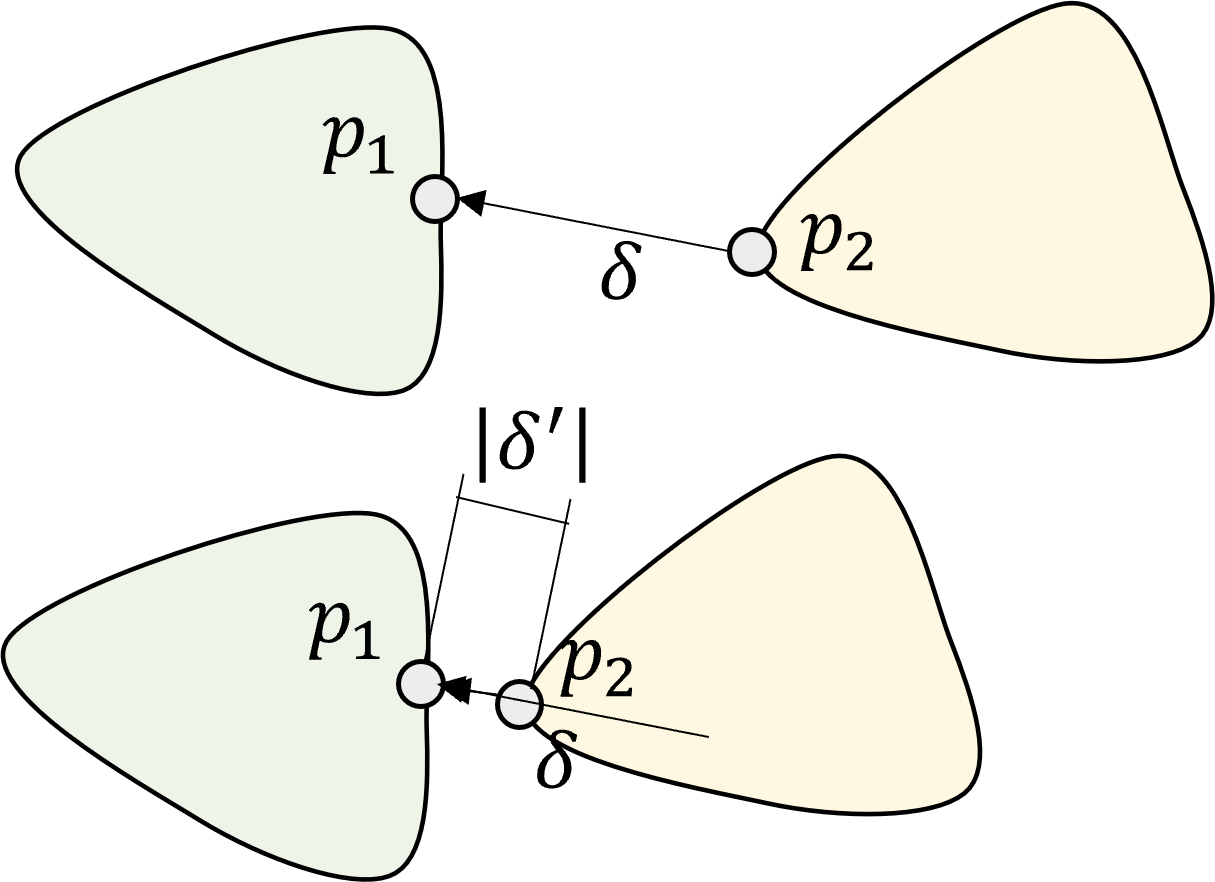}
    \caption{
    An illustration of distance manipulation for generating the collision dataset. We first place two objects (green and yellow)  far apart by sampling their poses. We then compute the minimum distance vector, $\delta$, and points $p_1$ and $p_2$ on this vector on each object. Then, we move one object  by $|\delta|-|\delta'|$ in the direction of $\delta$ to obtain a data point with the desired distance of $|\delta'|$. The desired distance $|\delta'|$ is sampled from a normal distribution.
    }
    \label{fig:dataset_distance}
\end{figure}

So, we devised a new strategy: after we sample poses with large enough bounds, we manipulate the distance between two objects to create collisions with different overlapping volumes. Figure~\ref{fig:dataset_distance} demonstrates our strategy. More concretely, we first compute the shortest-distance vector, $\delta$, between two objects, and then translate one of them along the direction of $\delta$. This way, for a given object pose pair, we can get collision data points at different overlapping volumes by manipulating the magnitude of $\delta$. 
In our implementation, we sample the target magnitude $|\delta'|$ from a normal distribution with zero mean and 0.020m standard deviation, and take an absolute value.

Using this dataset, \locc~is trained end-to-end with binary cross-entropy loss, with a
regularization term to prevent the shape encoder from overfitting. The loss function of \locc~is given by
\begin{align*}
    \sum_{(d,y)\in \mathcal{D}}&{BCE(f^{CP}(f^{SE}(d)),y) + \alpha |f^{SE}(d)|^2}\ ,
\end{align*}
where $\mathcal{D}=\{d^{(i)},y^{(i)}\}_{i=1}^{n}$ is dataset, $d^{(i)}=\{x_1^{(i)},x_2^{(i)},q_1^{(i)},q_2^{(i)}\}$ is the input, $f^{SE}$ is the shape encoder,
$f^{CP}$ is the collision predictor, and $BCE(\hat{y},y)$ is the binary cross entropy between the prediction
and label $y$.

During training, we apply a data augmentation scheme to make \locc~robust to different orientations. Concretely, suppose we have a data point $\{x_1^{(i)},x_2^{(i)},q_1^{(i)},q_2^{(i)},y^{(i)}\}$. We first randomly sample a rotation matrix $R$ and then apply $R$ to both shapes, and $R^{-1}$ to both poses to make a new data point $\{R \cdot x_1^{(i)},R \cdot x_2^{(i)},R^{-1}\circ q_1^{(i)},R^{-1}\circ q_2^{(i)},y^{(i)}\}$, where $R\cdot x$ denotes rotating the shape by $R$, and $R\circ q$ is applying rotation $R$ to the rotational part of pose $q$. Note that we can keep the same label even after these operations because $\{x, q \}$ and $\{ R\cdot x, R^{-1} \circ q \}$ occupy the same volume in the 3D space.

\section{Experiments}

\begin{figure*}[t]
    \centering
     \begin{subfigure}[b]{0.3\textwidth}
         \centering
         \includegraphics[width=\textwidth]{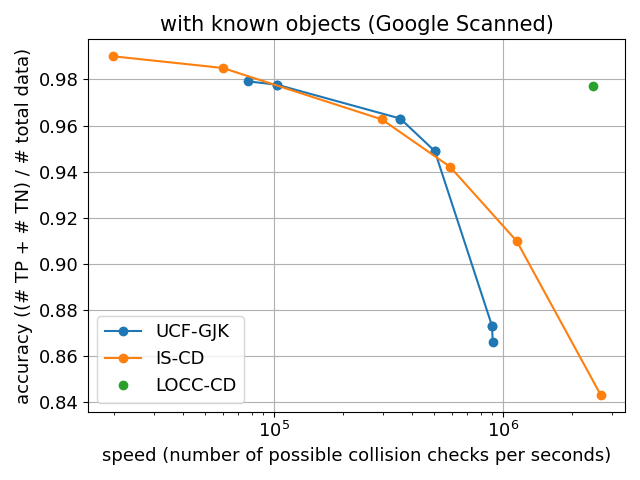}
         \caption{}
         \label{fig:acc_speed_plot_known}
     \end{subfigure}
     \begin{subfigure}[b]{0.3\textwidth}
         \centering
         \includegraphics[width=\textwidth]{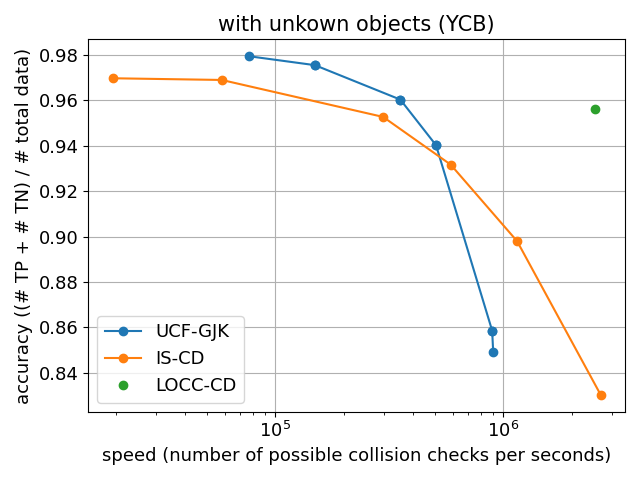}
         \caption{}
         \label{fig:acc_speed_plot_unknown}
     \end{subfigure}
      \begin{subfigure}[b]{0.3\textwidth}
         \centering
         \includegraphics[width=\textwidth]{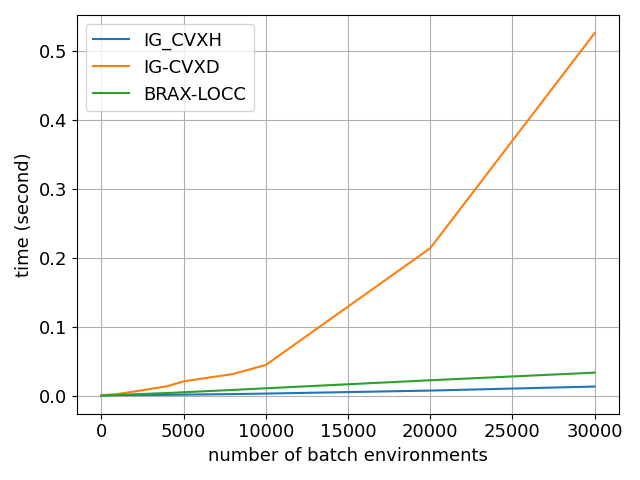}
         \caption{}
         \label{fig:simulation_efficiency}
     \end{subfigure}
    \caption{Accuracy-speed plots for (a) known and (b) unknown object sets. \# TP stands for number of true positive data pairs, and \# TN stands for number of true negative data pairs. (c) Plot of computational time it takes to simulate $\Delta t$ vs number of environments being simulated in parallel. }
    \label{fig:acc_speed_plot}
\end{figure*}

\subsection{Details of training}
For the hyperparameters of \locc~, we set $K$ to 1500, and uniformly sample 1500 points from the surface of the object mesh. In the shape encoding, we have $M=6, H=256$, $F=64$. More details about the architecture and the training hyperperameters are included in the supplementary material.

We use Google Scanned Object (GSO)~\cite{downs2022google} as our data that has 1030 object meshes most of which are non-convex. We use the meshes to generate point clouds and define the center of the mesh as the center of AABB. Unless otherwise mentioned, we generate a total of 230 million data points with 230K object pairs with the strategy in Sec. \ref{sec:dataset_preparation_and_training}.
For training, the learning rate is set to 0.001 and the batch size to 32. The Adam~\cite{kingma2014adam} optimizer is used with $\alpha$ set to 0.5 in the loss. See the appendix for more information on network parameters.

For testing, we use two distinct test sets. The first test set, called \emph{known object set}, uses the same set of objects as the training set, but different poses and object pairs. The total number of object pairs in the training set is 230K, and there are approximately 1 million possible object pairs. The second test set, called \emph{unknown object set}, consists of 30 novel objects chosen from YCB object set~\cite{calli2015ycb} that were not used during training. In generating the poses for test sets, we use uniform sampling rather than the method from section \ref{sec:dataset_preparation_and_training} to better characterize the poses that will be encountered during simulation.

\subsection{Results}
We wish to validate the following claims through our experiments:
 \begin{itemize}
     \item \textit{Claim 1 (computational efficiency of~\locc)} For non-convex objects, CD with \locc~is more computationally efficient than analytical CD algorithms such as GJK and CD based on implicit shape representation such as~\cite{driess2022learning}.
     \item \textit{Claim 2 (data efficiency)} \locc~is more data efficient than the standard OCN that learns the global object shapes to determine collision such as~\cite{danielczuk2021object}.
     \item \textit{Claim 3 (computational efficiency and generality in physics simulation)} For non-convex objects, physics simulation with \loccsim, which integrates Brax, \locc, and the contact resolution algorithm from~\cite{son2020sim}, is faster and has higher fidelity than IsaacGym, and more general than Brax which can only simulate convex objects.
 \end{itemize}

To support \textit{Claim 1 (computational efficiency)}, we use two baselines. The first is the UCF version of GJK written in JAX so that it can utilize XLA to run faster than the standard GJK on a GPU. We denote this as \ucfgjk. The second is CD based on implicit surface (IS), denoted \iscd, which determines collision based on a set of query points as in~\cite{driess2022learning}. 


\begin{figure}[htb]
     \centering
     \includegraphics[width=0.49\textwidth]{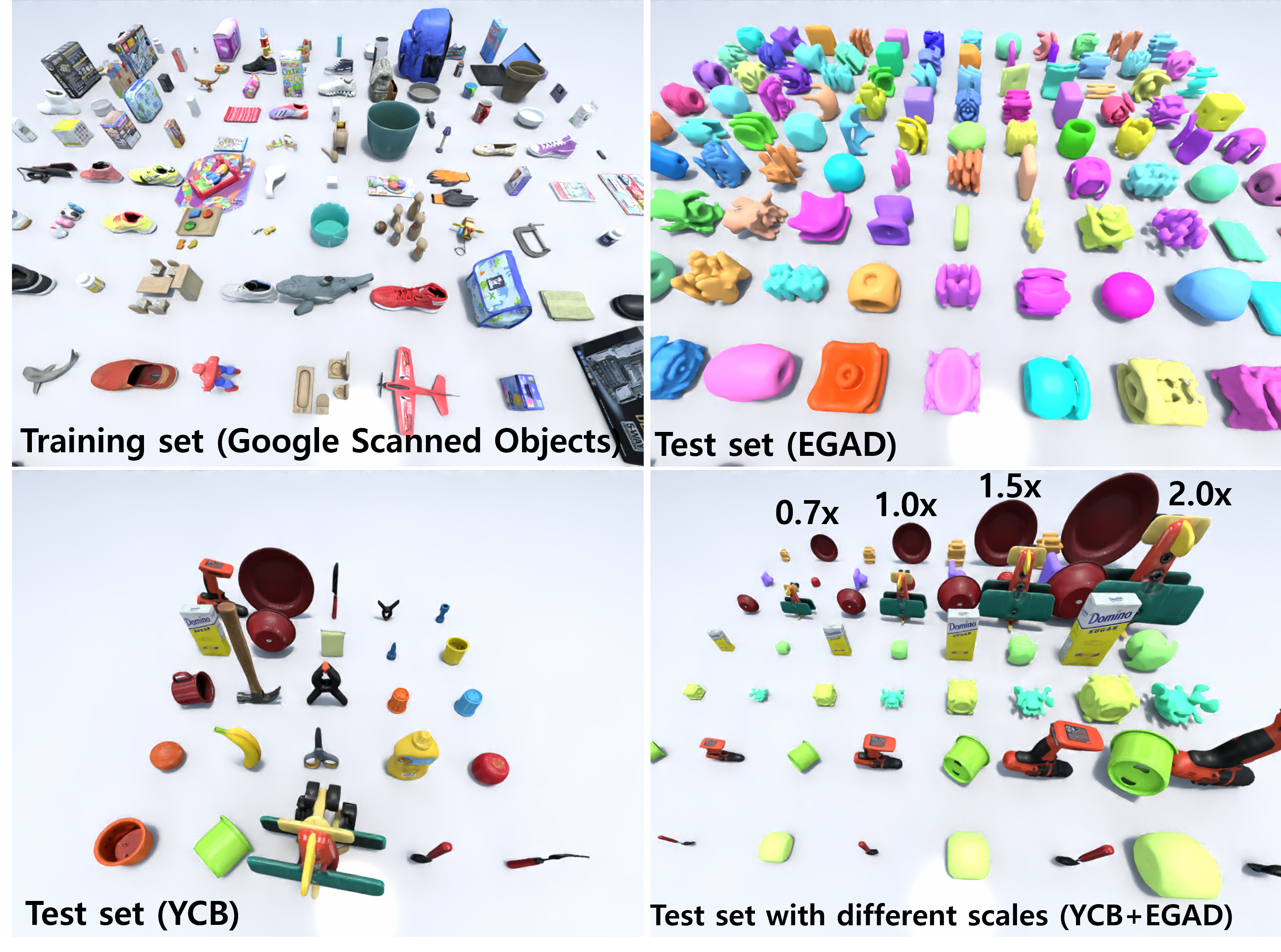}
     \caption{This figure illustrates a selection of objects utilized during training and testing phases. 
     The upper left quadrant features Google Scanned Objects~\cite{downs2022google} employed for the training process. During testing, objects from both the YCB~\cite{calli2015ycb} (bottom left) and EGAD~\cite{morrison2020egad} (upper right) datasets are implemented to demonstrate the generalization capability of our \locc model. Further, we assess \locc's adaptability to varying object scales by adjusting the YCB and EGAD objects by factors of 0.7x, 1.0x, 1.5x, and 2.0x (displayed in the bottom right quadrant). The result is a consistent collision accuracy exceeding 95\% across all test cases.
     }
     \label{fig:object_set}
\end{figure}

\begin{figure}[htb]
     \centering
     \includegraphics[width=0.45\textwidth]{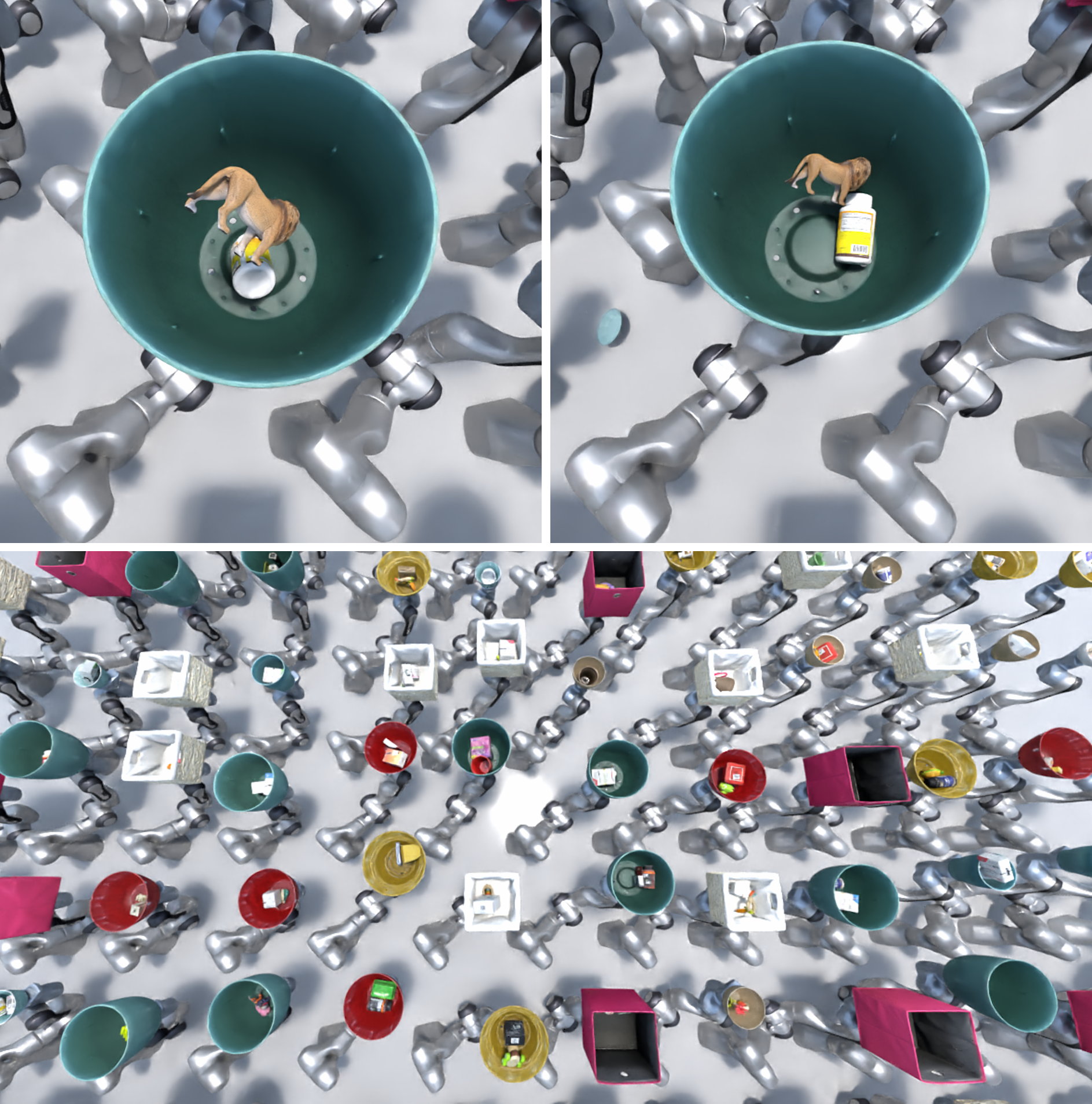}
     \caption{Object shaking simulation task comparing IsaacGym and \loccsim.}
     \label{fig:experiment_env}
\end{figure}

Unlike \locc~whose online computation time is fixed and its 
accuracy depends on the off-line dataset,
\ucfgjk~and \iscd~must trade off their computation time for
higher accuracy by increasing the number of elements in decomposition or the number of query points. Therefore, we use the accuracy versus speed plot to evaluate their performances, where the speed is  defined as the number of possible collision checks in a second, and accuracy is defined as the number of true positive and true negative predictions  divided by the total number of predictions. For \ucfgjk, there are  multiple hyperparameters, so we tested several hyperparameters and  report the one that attains the best performance. 

Figures \ref{fig:acc_speed_plot_known} and \ref{fig:acc_speed_plot_unknown} show the results for the known and unknown object test sets and respectively. In Figure \ref{fig:acc_speed_plot_known}, the result shows that \locc~achieves the accuracy of about 98\% and 96\% for known and unknown object sets respectively. This is comparable
to the best accuracy attained by \ucfgjk~and \iscd, which are 98.5\% and 98\% for known and unknown object sets respectively. However, we can see that \locc~achieves its best accuracy at a speed at least 10 times faster than both of these methods. This illustrates the advantage of \locc: by directly predicting collisions using a NN, it attains a faster computation time than \ucfgjk~and \iscd with comparable accuracy. 

Pursuing our investigation into the generalization capability of our method, we conducted supplementary experiments that involved training and testing datasets with significantly varied shapes. In these experiments, our model was trained solely on the GSO dataset, and then tested on two distinct datasets: EGAD~\cite{morrison2020egad}, a compilation of unusually-shaped objects created to evaluate the robustness of grasping algorithms, and the YCB dataset~\cite{calli2015ycb}, which we adjusted for scale (0.7x, 1.0x, 1.5x, and 2.0x). Representative samples from both the training and testing stages can be found in figure \ref{fig:object_set}.
Remarkably, our method consistently delivered prediction accuracy of at least 95\% across all the testing datasets. Such results substantiate the robust adaptability of \locc.


To support \textit{Claim 2 (data efficiency)}, we adapt SceneCollisionNet~\cite{danielczuk21icra} to object-object collision detection. This method does locality and must learn object shapes. We denote this method as \ocnglobal.
The shape encoder of \ocnglobal~is the same as \locc, except that the features are globally average-pooled before the de-convolution network in the collision predictor to extract the global features. To check the data efficiency, we train \ocnglobal~and \locc~with a varying number of object shapes, and for each number of objects, we collect 0.1 million object pairs and poses and test \locc~and \ocnglobal~on the novel object set. Table \ref{tab:global_local_table} shows the result.

When the number of training objects is 5 or 20, \locc~has 7-10\% higher accuracy than \ocnglobal~on average. As you increase the number of training objects, the gap gets smaller, indicating that although \ocnglobal~can eventually learn to generalize to novel objects, \locc~does so with a smaller number of object shapes. This supports our intuition shown in Figure~\ref{fig:locality_intuition}: \locc~outperforms \ocnglobal~because while \locc~can just encode the local geometric patterns at contacts which are similar across different objects, \ocnglobal~must learn to encode global object shapes for which there is much more variation.

\begin{table}[h!]
\centering
\begin{tabular}{|l ||c c c c|} 
\hline
\multirow{2}{*}{Methods} & \multicolumn{4}{c|}{\# objects} \\
& 5 &20 & 50 & 100 \\
\hline
\ocnglobal & 0.67$\pm$0.02 &0.83$\pm$0.03 & 0.92$\pm$0.02 & 0.93$\pm$0.01 \\
\locc~ & 0.77$\pm$0.02 &0.90$\pm$0.02 & 0.93$\pm$0.01 & 0.93$\pm$0.01 \\
\hline
\end{tabular}
\caption{Accuracy with varying number of objects in the training data.}
\label{tab:global_local_table}
\end{table}

Finally, to validate  \textit{Claim 3 (computational efficiency in physics simulation)}, we compare \loccsim~with IsaacGym on non-convex object simulation. We simulate 200 distinct environments with 3 objects, one of which is a bowl, and the rest are randomly selected from the Google Scanned Object set. The objects get dropped into the bowl, and the bowls get continuously shaken to create contacts. The example scenes are shown in Figure~\ref{fig:experiment_env}, and its video is included in the supplementary material. We could not compare \loccsim~to Brax because it is limited to convex shapes.  We set the size of the simulation time step, $\Delta t$, to 0.01/4 seconds, with 4 sub-steps.

\begin{figure}
    \centering
    \includegraphics[width=0.48\textwidth]{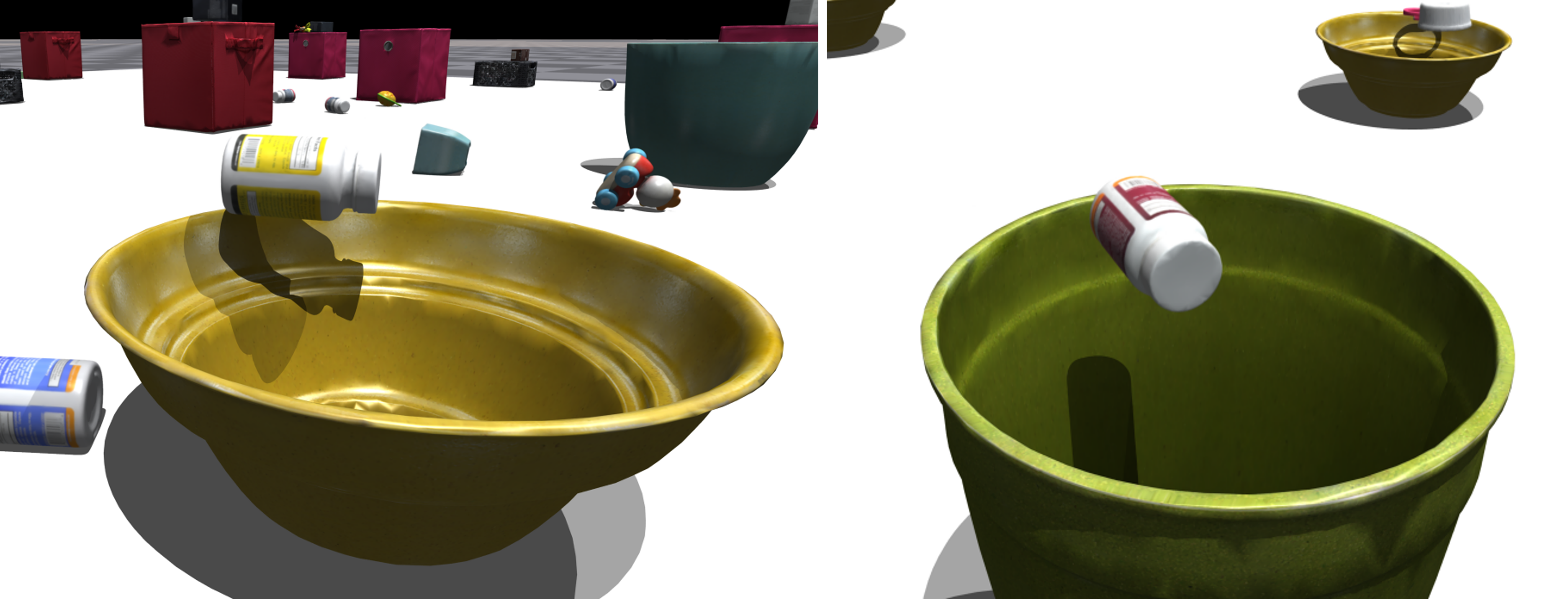}
    \caption{Example of inaccurate simulation behavior in~\igcvxh~due to a convex hull approximation of objects. The objects should fall into the bowls but are ``floating'' in the air because the bowls have been convexified}
    \label{fig:bowl_task_cvxhull}
\end{figure}

We measure the simulation speed without rendering. As Figure \ref{fig:simulation_efficiency} shows, \loccsim~is more than 10 times faster than~\igcvxd, which uses GJK and V-HACD, when we are simulating 30,000 environments. \loccsim~is slightly slower than \igcvxh, which uses a convex hull approximation of the object shapes. However, \igcvxh~has significant issues in simulating non-convex shapes, as demonstrated in Figure~\ref{fig:bowl_task_cvxhull}.


\newcommand{\minsd}{min-$|$sd$|$}
To measure this quantitatively, we compute the absolute sign distance among all the objects whenever a contact arises, and take the minimum value. We denote this as \minsd. This quantity measures the penetration depth if the objects penetrate each other at the contact or the distance between two objects when the contact is falsely detected. For an ideal simulator, this value would be zero at every contact.  
We use default parameters for GJK in IsaacGym. The results are shown in Table~\ref{tab:minsd stat}. 

As the table indicates, \loccsim~has the \minsd~value closest to zero among all the baselines, indicating that it has the least penetration depth and the least 
number of false positives. This is because \igcvxd~and
\igcvxh~inevitably lose their accuracy due to shape approximation, while the accuracy of \locc~depends solely on the quality and quantity of the offline training data.

\begin{table}[h!]
\centering
\begin{tabular}{|l ||c c c|} 
\hline
\multirow{2}{*}{Simulator} & \multicolumn{3}{c|}{metric (unit: meter)} \\
& average & top-10\% average & maximum\\
\hline
\igcvxd & 0.0132 & 0.0302 & 0.0392 \\
\igcvxh & 0.0709 & 0.1266 & 0.1286 \\
\loccsim (ours) & 0.0077 & 0.0218 & 0.0338 \\
\hline
\end{tabular}
\caption{Statistics on 600 absolute signed distance between two objects. \textit{average} is the average of \minsd, \textit{top-10\% average} is the average of highest 10\%, and \textit{maximum} is the highest \minsd~out of 600.}
\label{tab:minsd stat}
\end{table}

\section{Conclusion}
We proposed a novel OCN, \locc, that compared to previous approaches~\cite{das2020learning,danielczuk2021object}, is more data or computationally efficient. We showed that compared to analytical contact detection algorithms, our approach achieves higher accuracy and speed
when it comes to simulating non-convex objects in multi-environment scenarios by making better use of GPU resources. We integrated \locc~into the open-source physics engine, Brax, along with the contact resolution algorithm from~\cite{son2020sim} and showed that \loccsim~simulates non-convex objects while Brax cannot, and show that it outperforms IsaacGym in terms of speed and fidelity.

\bibliographystyle{plainnat}
\bibliography{ref}

\end{document}


\maketitle
\thispagestyle{empty}
\pagestyle{empty}

\subsection{Architecture Details}
Regarding the architecture of the shape encoder, first, point clouds are encoded to generate point-wise features with 3 layers of MLP with 256 neurons. With extracted point-wise features, we apply max-pooling by assigning points to each cell which is defined by AABB of the object and pre-defined voxel size. We use 6 for voxel size. After that, we have $6\times 6\times 6$ voxel-wise shape embedding with the size of 256 for each voxel. To make the final shape embedding mix global and local information, we apply 3D convolution layers. We have four layers of 3D-CNN which have 128 channels and filter of size (3,3,3), and for the first 3D-CNN layer we do not apply padding while the other three layers have. We keep features for every layer to be used as a skip connection when we apply deconvolution layers. Just before deconvolution layers, we apply max pooling to get global features of which size is 256. The 3D features before max pooling are fed into the deconvolution layers. We have 4 deconvolution layers with the same parameters in reverse order as 3D-CNN layers. Between each layer, we have a skip connection with concatenation. After deconvolution layers, we tile global features to match the size of the voxel size and concatenate them.
Finally, we apply one linear layer to produce a cell-wise feature with size 16, and it is our shape embedding.

These shape embeddings are used to predict collision in the prediction stage. We have two shape embeddings with a voxel grid and apply a transformation to them with the given pose of the objects. We then identify the intersection region of two AABB with additional margin $\epsilon={\sqrt{a_1^2+a_2^2+a_3^3}/2}$, where $a_1, a_2, a_3$ is the length of one cell of the counter object. And we gather features of voxels that are within the intersection region, and they are average-pooled for each object to be fed into the collision predictor. The collision predictor receives a pair of shape embedding and poses for each object. Then, shape embedding and pose for each object are concatenated to be fed into MLP. This MLP consists of 3 layers of 128 neurons followed by max-pooling across object pairs. This generates one vector of size 128, and we apply MLP of 3 layers of 128 neurons, followed by a linear layer of outputting scalar values. Finally, we apply sigmoid to get collision prediction values from 0 to 1.
across all layers including the shape encoder and collision predictor, we use the Relu activation function.

\subsection{Hyperparameters in table}


\begin{table}[h]
\centering
\begin{tabular}{|l |c |} 
\hline
\multicolumn{2}{|c|}{training} \\
\hline
machine & RTX A6000\\
learning rate & 1e-3\\
batch size & 32 \\
$\alpha$ in loss & 0.5 \\
optimizer & Adam \\
b1 for Adam & 0.9 \\
b2 for Adam & 0.999\\
eps for Adam & 1e-8 \\
\hline
\multicolumn{2}{|c|}{model} \\
\hline
voxel size &  6\\
width of MLP in shape encoder & 256 \\
\# layers of MLP in shape encoder & 4 \\
\# channels of 3D CNN in shape encoder & 128 \\
\# layers of 3D CNN in shape encoder & 4 \\
filter of 3D CNN in shape encoder & (3,3,3) \\
\# channels of 3D Deconvolution layers in shape encoder & 128 \\
\# layers of 3D Deconvolution layers in shape encoder & 4 \\
filter of 3D Deconvolution layers in shape encoder & (3,3,3) \\
width of MLP in collision predictor & 128 \\
\# layers of MLP in collision predictor & 6 \\
F & 16\\
\hline
\multicolumn{2}{|c|}{range of parameters in UCF-GJK} \\
\hline
maximum number of iteration & 4-9 \\
maximum number of decomposition & 2-16 \\
maximum number of triangles in one convex hull & 16-64 \\
\hline
\end{tabular}
\label{tab:hyperparameters}
\end{table}